Article

# Analysis of the Motion Sickness and the Lack of Comfort in Car Passengers

Estibaliz Asua [1,*] , Jon Gutiérrez-Zaballa [1] , Oscar Mata-Carballeira [2] , Jon Ander Ruiz [1] and Inés del Campo [1]

1   Department of Electricity and Electronics, Faculty of Science and Technology,
    University of the Basque Country UPV/EHU, 48940 Leioa, Spain; j.gutierrez@ehu.eus (J.G.-Z.);
    jonander.ruiz@ehu.eus (J.A.R.); ines.delcampo@ehu.eus (I.d.C.)
2   Ikerlan, S. Coop, Jose Maria Arizmendarrieta Pasealekua, 2, 20500 Arrasate-Mondragon, Spain;
    oscar.mata@ehu.eus
*   Correspondence: estibaliz.asua@ehu.eus

**Abstract:** Advanced driving assistance systems (ADAS) are primarily designed to increase driving safety and reduce traffic congestion without paying too much attention to passenger comfort or motion sickness. However, in view of autonomous cars, and taking into account that the lack of comfort and motion sickness increase in passengers, analysis from a comfort perspective is essential in the future car investigation. The aim of this work is to study in detail how passenger's comfort evaluation parameters vary depending on the driving style, car or road. The database used has been developed by compiling the accelerations suffered by passengers when three drivers cruise two different vehicles on different types of routes. In order to evaluate both comfort and motion sickness, first, the numerical values of the main comfort evaluation variables reported in the literature have been analyzed. Moreover, a complementary statistical analysis of probability density and a power spectral analysis are performed. Finally, quantitative results are compared with passenger qualitative feedback. The results show the high dependence of comfort evaluation variables' value with the road type. In addition, it has been demonstrated that the driving style and vehicle dynamics amplify or attenuate those values. Additionally, it has been demonstrated that contributions from longitudinal and lateral accelerations have a much greater effect in the lack of comfort than vertical ones. Finally, based on the concrete results obtained, a new experimental campaign is proposed.

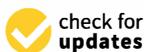

## 1. Introduction

Research on new generation of vehicles has been mainly focused on vehicle collision avoidance, lane keeping, coordination control or energy-efficiency with less attention paid to comfort [1–3]. Therefore, although the comfort-based strategies have not been the predominant control policies, they have to be considered in future vehicles, since when other activities are being carried out or no attention is paid to the trajectory both lack of comfort and motion sickness play a bigger role. Proof of this is the actual effort of vehicle manufactures to improve the vehicle comfort [4,5].

Quantifying the lack of comfort is a very challenging topic because it depends on the human perception and is affected by several factors. In [6], some types of human–vehicle interaction variables that influence the user's comfort are analyzed: human factors (as age and gender), environmental factors (as temperature, noise and pressure), spatial factors (as work-space, seat shape, etc.) and dynamic motion factors (usually named ride comfort).

In [7], Heibing et al. define ride comfort as the overall comfort and well-being of the vehicle's occupants during vehicle travel, whereas in [8], Heinz defines it as that part of the comfort construct that can be related to technically measurable or experienced vehicle movements, such as queasy static accelerations, shakiness and vibrations, as well as jerks. In [9], two terms have been used: discomfort (also named average discomfort), which is defined as a general feeling of not well-being, and motion sickness, that is associated with dizziness, fatigue and nausea.

The lack of comfort is associated with the frequency of the vibration and is directly proportional to the intensity of it. Furthermore, it has also been observed that increasing the time of exposure to vibration results in an increase of discomfort. It is known that low-frequency vibrations close to 1 Hz are transmitted throughout the body increasing malaise, while higher frequency vibrations are attenuated by the human body. It is also interesting to note that monotone continuous low-frequency vibrations increase fatigue, while transient vibrations produce stress [10]. Moreover, the human body responds differently to vibration frequencies depending on which body part and in which direction the force acts. How the frequencies affect humans depends on the proportions of a person's body and the type of frequency that affects the person [11].

While the fully automated vehicle (AV) is the future of the automotive industry, there is uncertainty regarding how the vehicle will be operating [12]. There is a strong correlation between the comfort and acceptance of autonomous vehicles [13,14]; so, if the factor affecting passengers' discomfort is known and the driving control strategy is accordingly improved, the riding comfort of passengers will be improved [15]. Research on the ride comfort of vehicles based on traditional drivers can provide a powerful reference for improving the algorithm design of the ride comfort of autonomous vehicles [15]. In [5], factors that could influence the comfort of autonomous vehicle motion are mentioned: acceleration, speed, quickness, vertical loadings, type of maneuver or driving style between others. It is also mentioned that given that these findings are based upon a limited number of research studies of various levels of realness, further evidence is required to confirm these factors.

Ride comfort analysis has been performed from different perspectives, being vehicle dynamics and the interior environment of the vehicle the most significant research lines. Thus, both the vehicle dynamics and the characteristics of the seat are also widely analyzed in several works, due to their influence in vibrations transmission [16]. Vehicle suspension designs have been proposed in [17] and experiments on seat structure suggest its influence on ride comfort measurements [18].

However, in addition, it seems obvious to conclude that both the characteristics of the road or the driving style affect passenger comfort. Few are the works dedicated to the analysis of the effect of the different characteristics of the journey in the lack of comfort. The works of [4,16,19] can be cited in this research line, where interactions between passenger comfort and pavement roughness or road are considered. The purpose of the works is to quantify the relationship between passenger comfort and the external environment.

Finally, as explained in [20], driving style plays an important role in vehicle energy management as well as safe driving. Driving style is understood as the way the driver operates the vehicle controls in the context of the driving scene and external conditions, such as time of the day, day of the week and weather, among other factors [20]. Some authors conduct research identifying drivers by classifying their braking patterns [21] and use CAN-bus signals information to evaluate their driving style [22]. In [23], a machine learning approach to identify aggressive and safe driving styles is proved by using discriminative features extracted from inertial signals. Driver's control of the brake, the throttle and the steering wheel results in low-frequency/high-magnitude disturbances that induce certain accelerations in the longitudinal and lateral directions with respect to a front facing passenger [24], and hence, driving style affects also passenger comfort. However, few works evaluate these activities from ride comfort perspective, among them [25–28] can be cited.

Automotive Seating Discomfort Questionnaire [29] and Automotive Seating Comfort Survey [30] are two official questionnaires related to comfort. A large number of studies on passenger comfort evaluation are based on questionnaires, subjects are asked to make subjective evaluation on riding comfort, and finally establish the relationship between passengers' subjective comfort and vehicle motion parameters [31–34].

So, although there are many works that address the ride comfort research line, as it has been already mentioned, the inter-relationship between different factors and scenarios and end-user comfort level needs to be further explored [5], even more so with a view to the self-driving car. Among the works that analyze the cause of the lack of comfort in different situations, or quantify how each of the causes mentioned above influence the lack of comfort of the user, the works of [12,35] can be cited. According to Griffin et al., vehicle ride quality, which is related to the dynamic response of the vehicle suspension and seating, concerns motion at frequencies greater than 1 Hz. The motion of vehicles at frequencies less than about 1 Hz arises from the profile of the road surface (for vertical vibration), travel around corners (for lateral acceleration) and acceleration and braking (for fore-and-aft motion). All these three are affected by vehicle speed and, in a different way, by driver behavior. On the other hand, in the recent work of [12], the feeling of comfort in terms of experienced motion sickness for the AV passengers is studied.

The objective of this work, in addition to completing the aforementioned investigations, is based on compiling the information and carrying out a complete study of those described possible different situations and factors that affect ride comfort, and quantify their effect on passenger comfort. The causes of the different disturbances have been analyzed from a passenger comfort perspective, evaluating signals that affect both motion sickness and the so-called general discomfort. With this purpose, we have selected a route that includes diverse road types. This route has been covered with two different cars and by three different drivers. An exhaustive experimental study of the importance of the road, the driving style and the vehicle itself in passenger comfort has been carried out. In addition, a ride comfort evaluation method based on a statistical analysis of probability density is presented, which completes the information of the power spectral analysis commonly used in ride comfort. Finally, in order to correlate the subjective feeling of comfort with quantitative data, four passengers have been subjected to these real experimental tests and they were asked to make subjective evaluation. Instead of using specific questionnaires [29,30], a generic one (but based on them) is used.

In summary, based on previous works, this paper presents a complete experimental methodology with an exhaustive method of analysis of comfort variables, which allows us to analyze how the different factors affect ride comfort as a whole, and which are also applicable to a more complex experimental campaign with the same purpose.

The remainder of this paper is organized as follows: Section 2 summarizes comfort evaluation methods whereas Section 3 explains the methodology carried out for the experimental campaign. In Section 4, a relative statistical analysis based on general comfort and motion sickness evaluation variables is presented. In order to extract more meaningful information, in Sections 5 and 6, a Probability Density Statistical analysis and a Power Spectral analysis are performed. In Section 7, passengers qualitative feedback is summarized. Once all the information is presented in previous Sections 4–7, in Section 8, the information is contrasted and analyzed. Finally, some concluding remarks and future work are summarized in Section 9.

## 2. Comfort Evaluation Methods

The comfort is usually evaluated, qualitatively, by using subjective rating tests [36] or, quantitatively, by using electrical accelerometers combined with the international standards [37,38]. The standards describe passenger comfort objectively, but cannot characterize human sensitivity differences for the same vibration, which often depend, for example, on the age, the gender and the weight of the passengers.

Two main standards are used in order to evaluate human exposure to whole-body vibration: the British Standard 6841 (BS 6841) [38] and the International Standard 2631 (ISO 2631) [37]. Both define methods for the measurement of vibrations and explain how to process measured data, to standardize quantified performance measures concerning health, perception, comfort and motion sickness. In this work, the ISO 2631 is used.

The quantified performance measures of ISO 2631 are based on frequency-weighted root mean square (RMS) computations of acceleration data as Equation (1) shows

$$a_{w_i} = \left(\frac{1}{T}\int_0^T (a_i(t), w_i)^2 \, dt\right)^{1/2} \quad (1)$$

where $(a_i(t), w_i)$ is the acceleration in the $i$-th axis weighted with $w_i$ filter in m/s² and $T$ denotes the time range of the measurement in seconds.

In this work, both motion sickness and general comfort are studied. Figure 1 summarizes the proposed ISO filters: $w_d$, the filter which evaluates the lack of general comfort for a seated passenger when accelerations are applied in longitudinal or lateral directions; $w_k$, the filter to assess the accelerations in vertical direction and $w_f$, the filter used in vertical direction but for motion sickness evaluation.

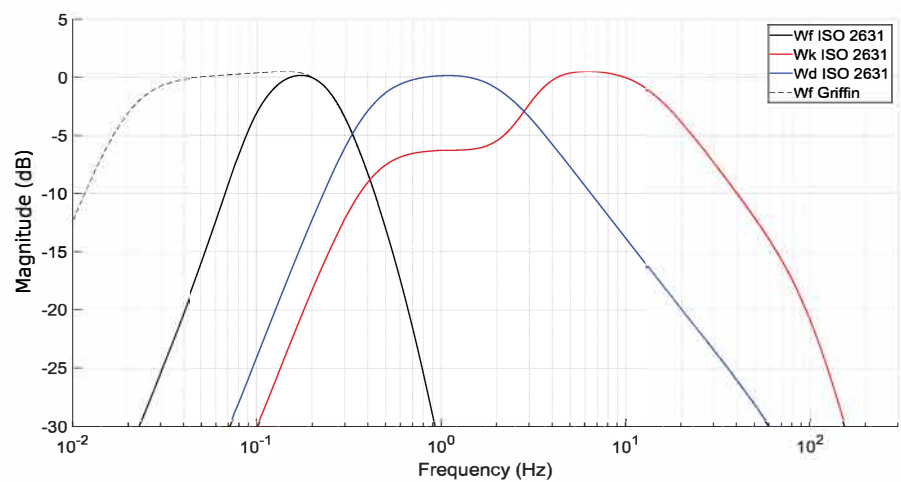

**Figure 1.** Filters proposed by ISO 2631 [37] and Griffin in [35,39–41].

As for general comfort, the $A_v$ parameter is presented. It accounts for the total weighted acceleration considering three orthogonal axes: vertical, lateral and horizontal axes. As it is expressed in Equation (2), the measured accelerations are first weighted with the corresponding filters and then multiplied by some factors which have to be taken into account as the perception of the acceleration depends not only on the place where the accelerometer is located but also on the direction of the measurement.

$$a_v = \left(k_x^2 a_{w_x}^2 + k_y^2 a_{w_y}^2 + k_z^2 a_{w_z}^2\right)^{1/2} \quad (2)$$

where $a_{w_i}$ is the weighted acceleration in the $i$-th axis in m/s² and $k_i$ are multiplying factors. Depending on its value, according to the ISO 2631 Standard, the ride comfort can be classified with different comfort grades as Table 1 indicates [42].

**Table 1.** Comfort index ($A_v$) and its relation with general discomfort [42].

| Range (m/s²) | Description |
| --- | --- |
| greater than 2.0 | extremely uncomfortable |
| 1.25–2.5 | very uncomfortable |
| 0.8–1.6 | uncomfortable |
| 0.5–1.0 | fairly uncomfortable |
| 0.315–0.63 | a little uncomfortable |
| less than 0.315 | not uncomfortable |

This standard can be used individually or in combination with acceleration and jerk peaks. In [43], a fore-and-aft and lateral ride comfort is evaluated using a methodology based on acceleration thresholds. A too high value of acceleration or jerk can cause discomfort because the passenger will find it difficult to maintain posture. Limit values vary among studies. In [44], a maximum acceleration value of 1.47 m/s$^2$ is defined, whereas in [9], the threshold is set at 2 m/s$^2$, as it is argued that, since passengers are seated when traveling in an automobile, the limit should be higher than the limit for trains and buses.

Motion Sickness Dose Value (*MSDV*) is a measure of the likelihood of nausea. The way to evaluate it is also defined in ISO 2631 Standard by a $w_f$ filter (see Figure 1) and Equation (3), where $(a_i(t), w_f)$ is the acceleration in the *i*-th axis weighted with $w_f$ filter in m/s$^2$. ISO 2631 only provides guidelines for the interpretation of the *MSDV* in the vertical direction, $MSDV_z$. However, in [35,39–41], Griffin justified that this variable may not be optimum for the prediction of motion sickness. Moreover, he experimentally demonstrated that motion sickness dose values are similar in longitudinal and lateral axes when calculated using the frequency weighting in current standards and that, at frequencies less than 0.1 Hz, fore-and-aft acceleration is greater than lateral acceleration. Based on those studies, he proposed an alternative to the $w_f$ ISO standard filter. This filter is represented in Figure 1 as $w_{f\ Griffin}$. Moreover, Forstberg provided an interpretation of the *MSDV* for the lateral direction in [6]. So, in this paper, *MSDV* in all directions is considered in order to analyze the effect on the passenger of each of the axes, and thus conclude if they could be combined as in the case of the $a_v$ variable or if, on the contrary, some contributions could be neglected.

$$MSDV_i = \left( \int_0^T (a_i(t), w_f)^2 \, dt \right)^{1/2} \tag{3}$$

Power Spectral Density (PSD) of the acceleration data could be another way of analyzing the lack of comfort. It has been mainly used when evaluating vehicle dynamics and suspensions as it is done in [45] for motorcycles, in [46] for rail vehicles, and in [47] for tractors.

PSD represents the power content across the frequency spectrum and so it can be very useful in order to determine the power distribution at certain frequencies [48]. One of the main methods used to estimate the PSD is Welch's Method [49]. Furthermore, in order to precisely analyze the details of the journey, time content can be added to the PSD, resulting in what it is known as a spectrogram [50].

Briefly, Welch's method is based on dividing the given data sequence $x[0], x[1], \ldots, x[N-1]$, where $N$ is the number of acquired points, into $K$ segments of $M$ points with a shift of $S$ between segments.

$$\text{Segment 1}: x[0], x[1], \ldots, x[M-1]$$

$$\text{Segment 2}: x[S], x[S+1], \ldots, x[M+S-1]$$

$$\text{Segment } K: x[N-M], x[N-M+1], \ldots, x[N-1]$$

For each segment, a windowed Discrete Fourier Transform (DFT) is computed at some frequency $\nu = i/M$ with $-(M/2 - 1) \leq i \leq M/2$:

$$X_k(\nu) = \sum_m x[m] w[m] exp(-j2\pi\nu m) \tag{4}$$

where $m$ varies from $(k-1)S$ to $M + (k-1)S - 1$ and $w[m]$ is the window function (Hann, Hamming, Blackman or Kaiser–Bessel, for example).

Then, for each segment, the modified periodogram value is generated from the DFT:

$$P_k(\nu) = \frac{1}{W} |X_k(\nu)|^2 \tag{5}$$

where

$$W = \sum_{m=0}^{M} w^2[m] \tag{6}$$

Finally, the periodogram is averaged in order to obtain Welch's estimate of the PSD:

$$S_x(\nu) = \frac{1}{K} \sum_{k=1}^{K} P_k(\nu) \qquad (7)$$

## 3. Methodology

In this section, the methodology carried out for the experimental measurements is explained. As it has been already mentioned, the selected route includes different types of roads which have been covered with two cars and three drivers. In addition, passengers have been questioned to assess their comfort in different situations.

### 3.1. Device for Data Acquisition

Among all the possibilities analyzed to choose a suitable measuring device, we consider the smartphone the most appropriate one, because it is non-intrusive and car-independent, in order to allow drivers a naturalistic way of driving [51,52]. Furthermore, it has to be taken into account that the device has to be user-friendly, so as to have a suitable platform for massive data collection. Attending to sample time, and considering the filters shown in Figure 1, we decided that the device has to be capable of sampling signals up to 25 Hz, being this frequency above which all filters attenuate the signals by at least half.

Finally, a ZTE-Blade-A452 Smartphone with Phyphox app installed has been used [53], which allows access to the accelerometers of the smartphone with a maximum sampling rate of 100 Hz. A screenshot of the application during data recollection is included in Figure 2a.

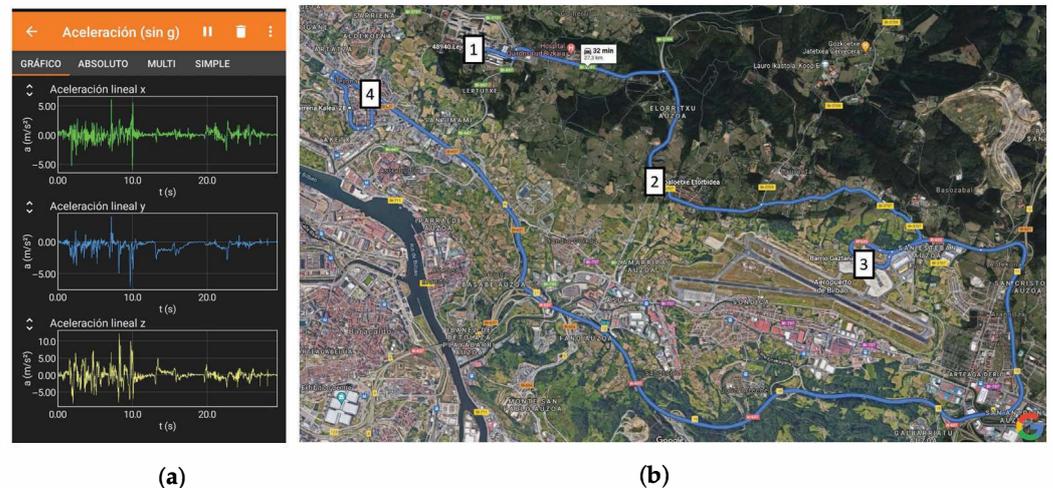

**Figure 2.** Selected data acquisition device and journey. (**a**) An example of a screenshot during data acquisition [53]; (**b**) selected 26-km journey near Bilbao, in the north of Spain.

### 3.2. Journey, Driver and Car Selection

The traversed route is shown in Figure 2b. The route covers a total of 26 km near Bilbao, in the north of Spain. It has been selected bearing in mind that it should contain representative and diverse sections, so as to asses the accelerations suffered by the passengers. It can be divided into 4 distinguishable sections:

- Section 1 (S1): From point 1 to point 2 (3.7 km). An interurban road with altitude changes (it first goes up and then down), some smooth curves and a couple of 90 degree turns. The average speed is 61.36 km/h;
- Section 2 (S2): From point 2 to point 3 (5.2 km). An interurban section with a remarkable change in height that includes lots of curves. The average speed is 44.60 km/h;
- Section 3 (S3): From point 3 to point 4 (15.4 km). A flat highway with overtaking and lane change possibility. The average speed is 80.72 km/h;

- Section 4 (S4): From point 4 to point 4 (2.2 km). A circular urban section with zebra crossings, roundabouts, traffic lights and bumps. The average speed is 30.35 km/h.

Figure 3 displays longitudinal, lateral and vertical accelerations experienced by the passengers across the representative journey. As it can be seen, there are meaningful differences among sections. For example, while section S3 (which corresponds to a highway) exhibits low accelerations in both longitudinal and lateral axes, S2 (a section with significant height differences) and S4 (an urban section) exhibit higher frequency accelerations, which are also higher in magnitude. Additionally, although the differences are smaller in the vertical axis, there are certain situations that show great values in magnitude. It has been verified that those situations correspond to zones where there are bumps.

The aforementioned route has been covered by three different drivers and two different vehicles, so it has been completed six times. The objective of incorporating variability into the study is to analyze the effect on passenger comfort of both driving style, which could vary depending on the gender and age of the drivers, and vehicle dynamics, which can be influenced by the type and age of the cars. Drivers are identified as D1 (40/50-year-old man), D2 (20/30-year-old woman) and D3 (20/30-year-old man), whereas vehicles are designated as C1 (crossover type, Nissan Quashqay) and C2 (sedan type, Opel Vectra).

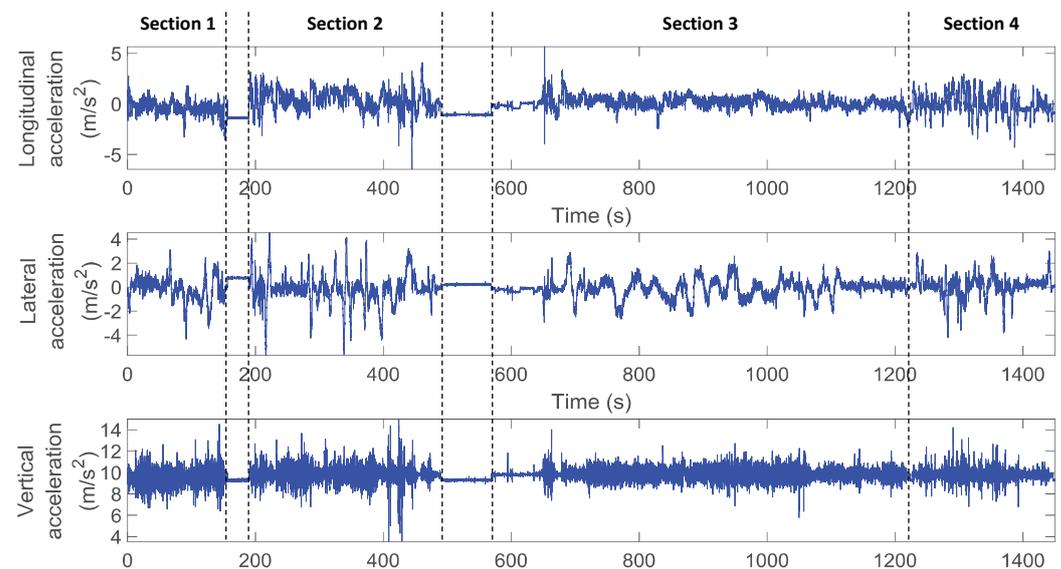

**Figure 3.** Longitudinal, lateral and vertical acceleration experienced by the passenger across the representative journey.

The differences in the driving style are shown in Figure 4, a representative graph where longitudinal, lateral and vertical accelerations for the same car being driven by two drivers completing Section 2 are depicted. In fact, one driver (blue) covers the journey faster (300 s) and generates more brusque maneuvers (especially in lateral axis) than the other one (red), who has a more conservative driving style and needs 100 s more to complete the section. Concerning vehicle differences, Figure 5 displays the data collected during the journey driven by the same driver and using both cars. Differences can be observed in both lateral and longitudinal directions, while the vertical accelerations remain similar.

It is expected that the disparity in road characteristics, driving style and vehicle dynamics will impact differently on passenger comfort. In the following sections, a comprehensive analysis of the causes associated with the lack of passenger's comfort will be conducted.

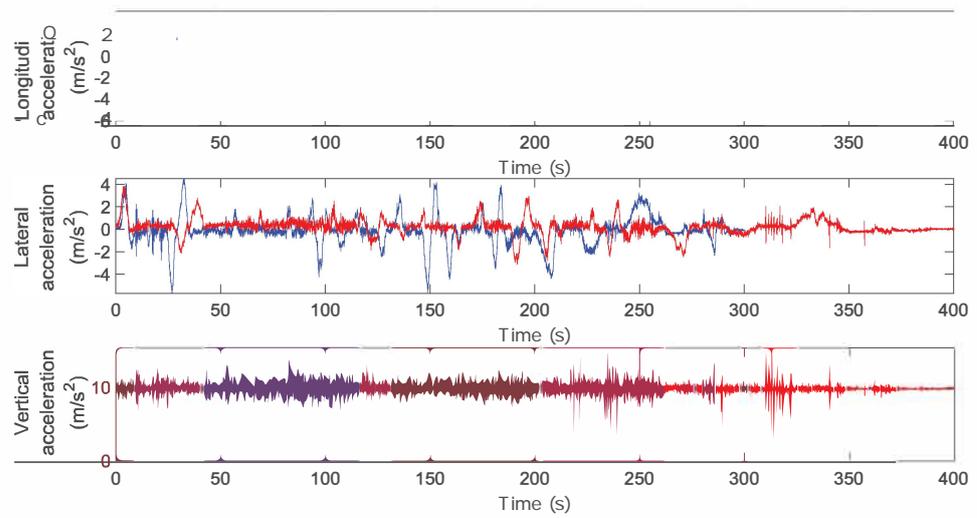

**Figure 4.** Longitudinal, lateral and vertical acceleration for the same car driven by two drivers to complete the Section 2, one driver (blue) being more aggressive than the other one (red).

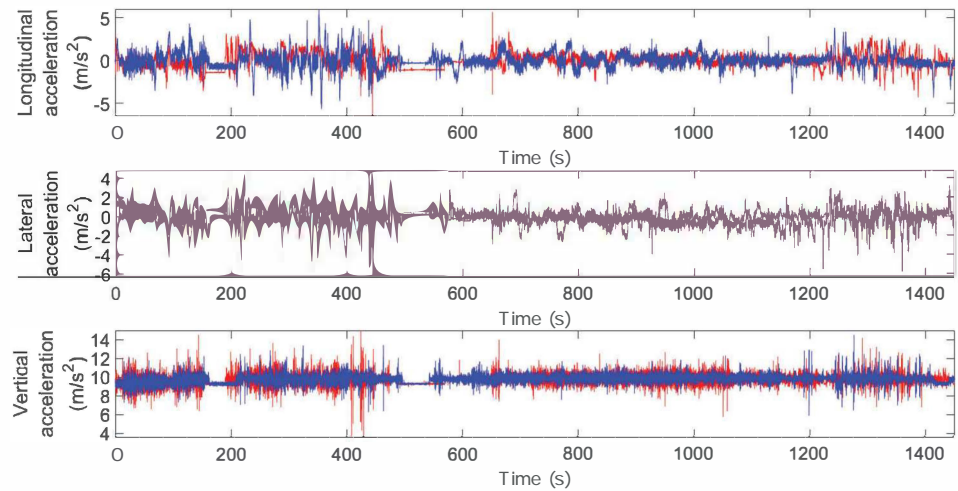

**Figure 5.** Longitudinal, lateral and vertical acceleration for the same driver driving both cars: Crossover (red) and Sedan (blue).

*3.3. Passenger Feedback*

During all data acquisitions, four passengers have been transported and, in order to asses their subjective feeling, all of them have given their feedback once the experimental campaign has finished. They have been requested to order the three drivers according to motion sickness and the lack of comfort felt during the route. They have also been required to remark the differences related to the car itself and, finally, to arrange the different sections according to both motion sickness and general comfort.

## 4. Relative Statistical Analysis

The goal of this section is to compare the values of the general comfort and motion sickness evaluation variables when all vehicles, drivers and sections are considered. Several variables have been calculated by axes in order to extract more meaningful conclusions. Their values have been calculated for each section, car and driver, that is, $A_{v_{D1C1S1}}$ represents the variable $A_v$ calculated using Equation (2) from acceleration measurements acquired along section S1 when car C1 has been driven by driver D1. Thus, for each variable 24 different values are calculated, and if driving style is analyzed, $A_v^{D1CiSj}$ is compared with $A_v^{D2CiSj}$ and $A_v^{D3CiSj}$ one by one for all $i$ and $j$ values.

*4.1. General Comfort Analysis*

The first part of the analysis focuses on frequencies higher than 1 Hz, that is, accelerations which are believed to cause what is known as general discomfort. As it has been previously explained in Section 1, both $A_v$ calculated by Equation (2) and $N_i$ could be used for this purpose. $N_i$ represents the times that acceleration data exceed a certain threshold considered to cause discomfort, which has been set to 2 m/s$^2$. Figure 6 shows a comparison between drivers, cars and sections when $A_v$, $N_x$, $N_y$ and $N_z$ are considered. The percentages of situations where car $i$ exceeds car $j$ (Figure 6a) or driver $i$ exceeds driver $j$ (Figure 6b) or section $i$ exceeds section $j$ (Figure 6c) for the comfort variables are shown. Each specific color represents each car or driver or section, and they are compared one by one. When the difference between the calculated variable is less than 5%, situations which have been considered as non-determinant, it is represented in beige. That is, in Figure 6b, D1 (red) is compared against D2 (blue) and D3 (green), and next, D2 (blue) is compared against D3 (green). If $A_v$ is analyzed, on the one hand, it can be concluded that in 60% of the situations D1 presents a higher $A_v$ than D2, in 10% a lower $A_v$ and, in the remaining cases (beige), it could be considered very similar values. On the other hand, D1 always presents higher values of $A_v$ than D3.

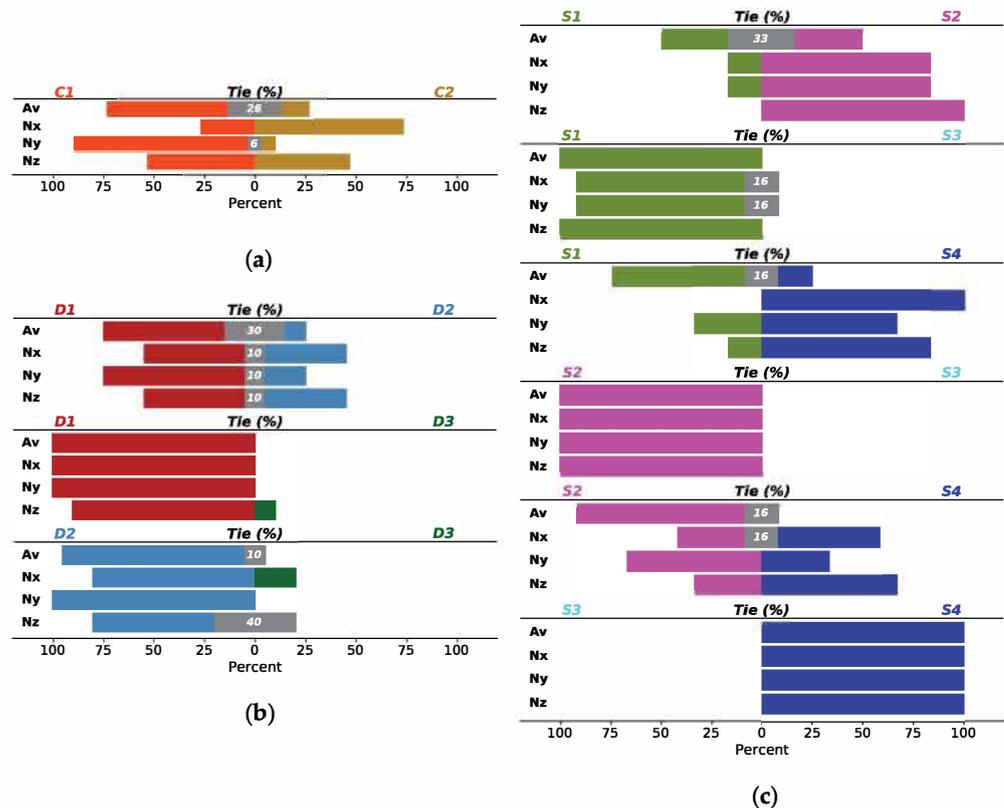

**Figure 6.** Comparison between drivers, cars and sections when $A_v$, $N_x$, $N_y$, $N_z$ are considered. The percentages of situations where car $i$ exceeds car $j$ (**a**); driver $i$ exceeds driver $j$ (**b**); and section $i$ exceeds section $j$ (**c**) for each particular variable.

Considering this figure, taking vehicle differences into account, when comparing C1 and C2 (Figure 6a), it can be concluded that C2 exceeds the threshold in the longitudinal axis ($N_x$) more frequently than C1 does, but in the lateral axis ($N_y$), just the opposite happens. Attending to $N_z$, we cannot conclude anything, since in half of the situations, C1 outperforms C2 and the other half vice versa. Concerning $A_v$, even though the greatest values are related to car C1, the differences between C1 and C2 are too small and almost all the recorded $A_v$ values belong to the same discomfort level. Furthermore, if the threshold to distinguish the two values increases, the percentage of times that C1 exceeds C2 decreases significantly (see Figure 7).

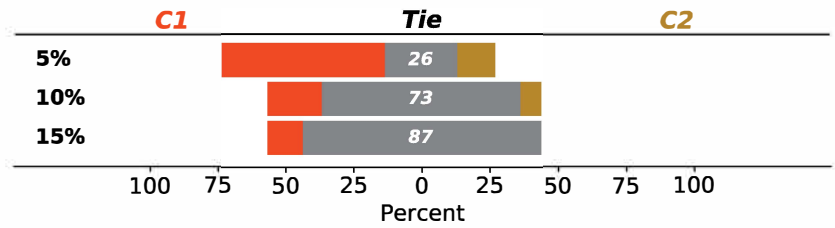

**Figure 7.** Differences in car comparison percentages for $A_v$ variable when several threshold values are selected in order to consider the difference between them significant.

On the other hand, concerning driver comparison (Figure 6a), differences can be seen in their driving style. As all variables show, it can be undoubtedly concluded that D3 is the one who drives most comfortably and D1 the least.

Finally, the results for the different sections are analyzed. As a consequence of the difference in their length, in order to compare them, $N_x$, $N_y$ and $N_z$ have been averaged in distance. Undoubtedly, S3 is the one in which the threshold is surpassed the least in all directions and $A_v$ is smallest, S1 is the second most comfortable, whereas S2 and S4 are the worst ones. S4 is less comfortable if longitudinal ($N_x$) and vertical ($N_z$) axes are considered, whereas S2 is less comfortable in lateral direction ($N_y$). The value of $A_v$ is similar for S1 and S2 sections, and the S4 section is more comfortable in this sense than the other sections.

*4.2. Motion Sickness Analysis*

Once general comfort has been analyzed, motion sickness has been examined. Figure 8 summarizes the results for *MSDV* values in the three directions calculated using Equation (3).

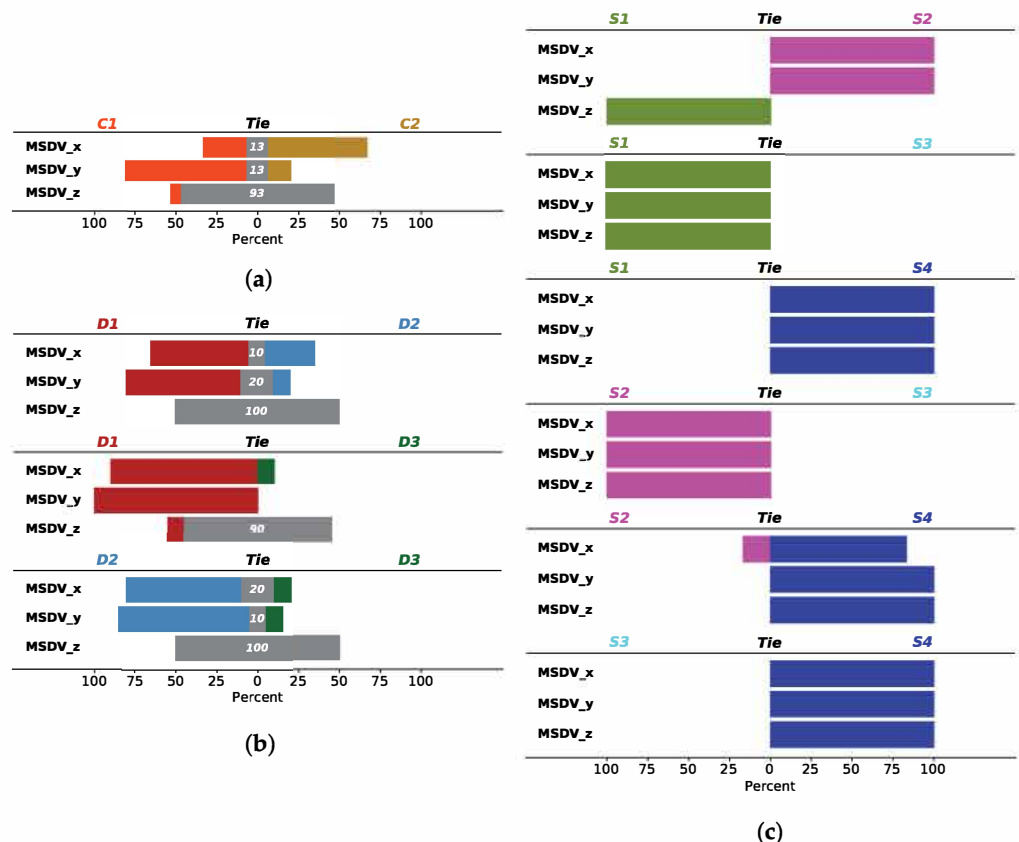

**Figure 8.** Comparison between drivers, cars and sections when $MSDV_x$, $MSDV_y$, $MSDV_z$ are considered. The percentages of situations where car *i* exceeds car *j* (**a**); driver *i* exceeds driver *j* (**b**); and section *i* exceeds section *j* (**c**) for each particular variable.

The comparison between cars coincide with the ones for the general comfort, as C1 is also more uncomfortable at low frequencies in the lateral axis, whereas C2 is less comfortable in the longitudinal one. Moreover, there are no meaningful differences in the vertical axis.

The comparison between drivers reveals that D3 is the one that produces the least motion sickness. In addition, there is also a huge difference between D2 and D1, being the latter the most dizzying one. However, if $MSDV_z$ is considered, there are not remarkable differences.

In order to compare sections, and due to the accumulative character of $MSDV$, it has been averaged considering the distance. Once again, S4 is the one with the highest values in all directions. Then, S2 can be considered as the second least comfortable in both lateral and longitudinal axes, but $MSDV_z$ is greater for S1 than S2. Finally, the most comfortable in all directions is S3.

## 5. Probability Density Statistical Analysis

The statistical analysis performed in the previous section has brought some overall conclusions about the drivers, the cars and the sections. However, in order to extract more meaningful information, next, a different statistical approach to several comfort parameters is provided. With this purpose, the kernel density estimation (KDE) is used. The KDE technique, unlike histogram, produces a smooth estimate of the probability density function (PDF) and is able to suggest multimodality [54]. It is useful to estimate the PDF of datasets which are difficult to be modeled by parametric density functions. The formal definition of KDE is a function defined by equation

$$f_h(x) = \frac{1}{nh} \sum_{i=1}^{n} K(\frac{x - x_i}{h}) \tag{8}$$

where $K(x)$ is called the kernel function that is generally a smooth, symmetric function such as a Gaussian, n number of samples, $x$ a given point and $h > 0$ the smoothing bandwidth that controls the amount of smoothing [54].

In this section, the variables have been calculated approximately every 10 s, that is, we have selected a window size of 1024 points which, taking into account the sampling frequency of the device, corresponds to 10.24 s.

It has been observed that $N_x$ and $N_y$ are scattered variables and, for a high number of windows, there are no situations in which the accelerations surpass the threshold; so, kernel density analysis makes no sense. On the other hand, as it has already been concluded, when $A_v$ is considered, in some of the cases, there are no significant differences among cars, drivers and sections, being its value very similar. Finally, when $MSDV$ variable's magnitude is analyzed, it can be concluded that its value in the vertical axis is much smaller than the longitudinal and lateral ones.

Thus, Figure 9b depicts the bivariate kernel density function of the $MSDV_x$ and $MSDV_y$ corresponding to the four sections and the three drivers (data collected with both vehicles have been considered). The color represents the probability density, from red (improbable) to pink (probable) through yellow, green and blue. In addition, the closer the lines, the more regular the conduction is.

As we already know, this figure shows that D1 produces more motion sickness than D2 and that both of them generate far higher values than D3. Besides, this analysis demonstrates that D3 presents a rather regular and comfortable driving style and that D1 can be considered to be a less regular driver. It is worth mentioning that, although D2 drives most of the time with lower $MSDV$ than D1, she shows an irregular driving comparing with D3, which increases her average $MSDV$.

With regards to the sections, the results are consistent with what has already been explained in the previous section: S3 and S1 are the ones with the least motion sickness and S2 and S4 the ones with the most. This analysis also demonstrates that all drivers exhibit a more spread kernel in the latter situations.

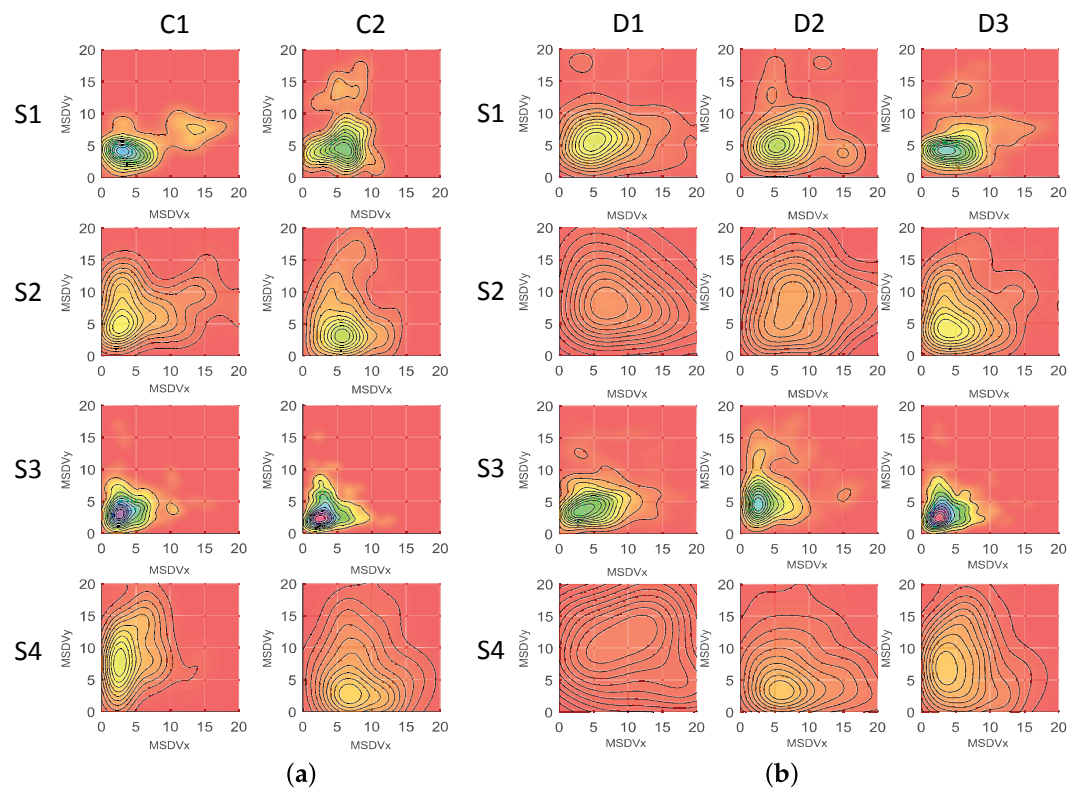

**Figure 9.** Bivariate kernel density function of the $MSDV_x$ and $MSDV_y$ for: (**a**) all sections (rows) and both vehicles (columns) when the driving style of D3 is considered; (**b**) all sections (rows) and all drivers (columns). The color represents the probability density, from red (improbable) to pink (probable) through yellow, green and blue. In addition, the closer the lines, the more regular the conduction is.

Figure 9a resumes the differences when vehicle analysis is considered. As it has already been demonstrated that driver D3 is the most regular one when the driving is considered, the figure focuses on his driving, but the conclusions can be extrapolated for the rest of the drivers. As can be seen, concerning the regular driving, during the most comfortable sections (S1 and S3), the driving seems to be more regular. Besides, car C1 presents greater values for $MSDV_y$ than C2 and smaller values for $MSDV_x$ than C1, especially in the less comfortable sections.

## 6. Power Spectral Analysis

With the aim of completing the thorough study of the lack of comfort and its causes, a frequency-domain analysis (power spectral analysis) of the accelerations suffered by the passengers in all directions has been carried out.

In order to illustrate the results we have obtained with this analysis, Figures 10 and 11 show the power accumulated by two different drivers (D2 and D3) along sections S1, S2 and S4 with the same car for all directions (S3 has not been included as there are no meaningful differences) and the power accumulated by the same driver (D1) along the same sections but with both cars, respectively.

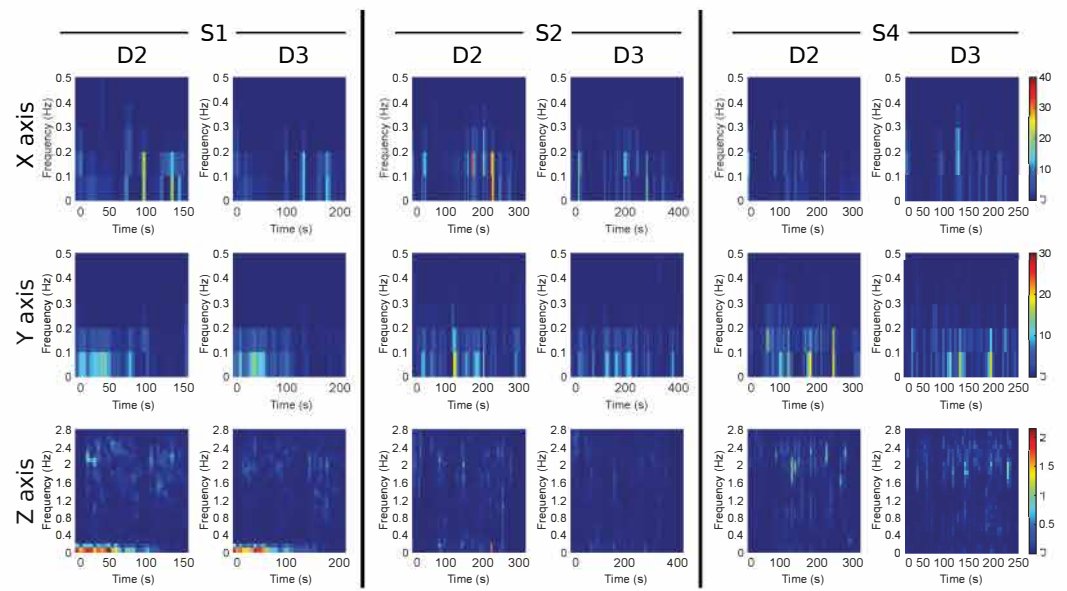

**Figure 10.** Spectrogram of the longitudinal (*x*), lateral (*y*) and vertical (*z*) axes (rows) of D2 and D3 drivers (columns) with C2 car when S1, S2 and S4 sections are considered (blocks).

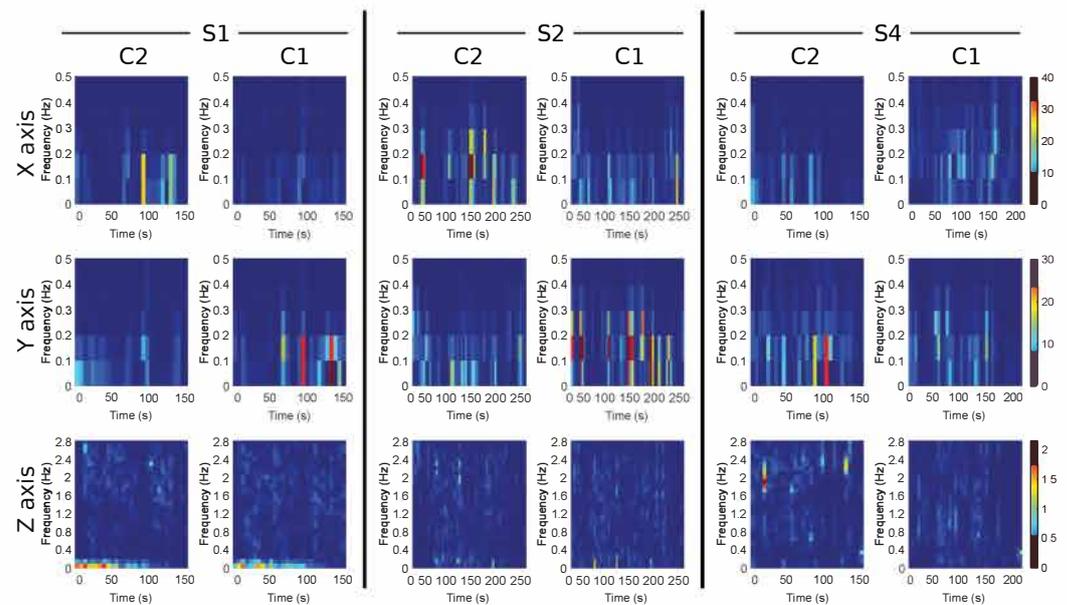

**Figure 11.** Spectrogram of the longitudinal (*x*), lateral (*y*) and vertical (*z*) axes (rows) of D1 driver with both cars (columns) when S1, S2 and S4 sections are considered (blocks).

According to the spectrograms, the frequencies of the accelerations that accumulate the most power in the lateral and longitudinal directions are in the 0–0.3 Hz range for every section, driver and car. In the vertical axis, there is also another frequency around which a relevant amount of power is condensed: 2 Hz (this phenomenon is especially visible in section S4). However, the power in this axis is much lower than in the other two (40/30 dB vs. 2 dB).

Moreover, attending to Figure 10, we can conclude that there is a common pattern during the travel both in longitudinal (*x*) axis and also lateral (*y*) axis, that is, the power tends to accumulate in the same parts of the path regardless of the driver (D2 and D3 are shown in this figure). If we pay attention to the longitudinal axis, in all sections, there are certain time instants for which both drivers have a higher power than for the rest of the section. The same occurs with the lateral (*y*) and vertical (*z*) axes, which is especially noticeable in S1, where, regardless of the driver, the greatest amount of power is gathered at very low frequencies, during the first one, and half a minute of the ride. However, the driver contributes to the magnitude of this power, which is greater for driver D2 in every case.

This effect is also shown when we introduce in the first column of Figure 11 the third driver's (D1) spectrogram when the same car is used. However, as we can conclude showing the same figure, the pattern changes when the other car is used. Anyway, the power for car C2 for the whole frequency range is greater comparing with the one for car C1.

## 7. Passenger Feedback

In order to correlate the results with the passenger comfort, as it has been reported in Section 3, their qualitative sensation has been analyzed.

Firstly, they have been asked to order the three drivers in terms of the motion sickness and the lack of comfort they generated. Unanimously, the feedback reported by the passengers consider driver D1 as the least comfortable, driver D3 as the most comfortable and driver D2 as an intermediate one. In detail, they also commented that driver D1 drove particularly uncomfortably in the longitudinal axis and that driver D3 drove more slowly and steadily.

When asked about the differences between the two cars, the passengers have not notified any meaningful ones. However, it is interesting to remark that drivers D2 and D3 expressed they felt significant differences when driving C2 car compared with their regular car. As a consequence of that feeling, they acknowledged they had been more conservative when driving C2 car.

Lastly, they have been asked to rank the sections attending the aforementioned parameters. In this case, there is not a clear conclusion. Although all passengers identified section S3 as the most comfortable one in every aspect (even though driver D3 tried to overtake in a certain moment which was considered to be an uncomfortable instant), there is no agreement about the least comfortable one: two passengers opted for section S4, whereas another two picked section S2. Thus, section S1 is the second most comfortable. Nonetheless, the general perception is that there are differences between section S4 and S2. The former is characterized for having uncomfortable events associated with bumps and fore-and-aft situations but, as it is ridden more slowly, it seems that motion sickness is greater in the latter.

## 8. Discussion

Based on the results obtained in Sections 4–7, an exhaustive analysis of them has been performed by studying the repercussion of the type of road, the car and the driving style on the lack of comfort of the passenger. Once the results and the feedback of the passengers have been analyzed, this section collects the main conclusions of the experimental campaign.

It is worth noting that in every direction, the frequency of the accelerations ranges from 0 Hz to 0.3 Hz (see Figures 10 and 11) for every car, driver and section and that in the vertical direction, there is a low-power characteristic frequency located at 2 Hz. Therefore, attending to general discomfort (filters focused on higher frequencies), none of the situations are specially uncomfortable. This correlates with calculated Av values, which according to Equation (2) are *'not or little comfortable'*. Passenger feedback confirms the mentioned result.

Additionally, all variables have been separately compared by axes for every car, driver and section. It has been verified that not only *N* but also *MSDV* have a significant smaller value for their vertical axis than for the other axes. This implies that the attention has to be paid to the lateral and longitudinal axes, which strongly agrees with the last work done by Griffin and Fortsberg [6,35,39–41] and conflicts with the ISO norm [37].

The study reveals that the type of road and its characteristics significantly influence the results of the comfort variables. On the one hand, both *N* and *MSDV* prove that, undoubtedly, section S3 is the most comfortable in every sense. S1 is considered the second most comfortable taking all variables except $MSDV_z$ into account. Paying attention to the spectrogram of the vertical axis of S1 in Figures 10 and 11, it is clear that much power is located at low frequencies and in the first part of the travel. After completing the journey again, it has been observed that, in the beginning of section S1, the pavement is irregular and rough which implies that, only taking $MSDV_z$ variable into account, S1 is more uncomfortable than S2.

On the other hand, S4 is the least comfortable section if longitudinal and vertical axes are considered (bigger $N_x$, $MSDV_x$ and $N_z$) and lateral low frequencies are evaluated (bigger $MSDV_y$), whereas S2 is less comfortable in the lateral direction when high frequencies are considered (bigger $N_y$). This matches with the section characteristics because, as it has been explained in Section 3, S3 is a headway, S2 has a lot of curves, whereas S4 has roundabouts, pedestrian crossings, traffic lights and bumps.

In addition, kernel analysis demonstrated that all the three drivers with both vehicles exhibit a more spread kernel in S2 and S4 (see Figure 9b), which means that the characteristics of the section themselves lead to more irregular drivings.

Furthermore, the spectrograms (Figure 10) depict a similar pattern in every section for every driver, which confirms that the features of the section are responsible for certain accelerations.

Passenger feedback coincides with the previous results as it characterizes the lack of comfort of the sections in the same way as the variables do. Moreover, with regards to section S1 and $MSDV_z$, no passenger did identify that section to be uncomfortable which shows that, indeed, the lateral and longitudinal axes contribute in a more substantial way than the vertical axis does.

Concerning driver comparison, it can be seen that driving style emphasizes the level of discomfort and fatigue that passengers suffer across the whole journey. Thus, as all variables show, it can be undoubtedly concluded that D3 is the one who drives most comfortably. On the other side, D1 is the least comfortable one, while D2 can be characterized as an intermediate one. Moreover, the different kernel density surfaces constructed from driving data represent the drivers' preferred driving style and show that D3 has the most regular driving style whereas D1 has the most irregular one. D2 drives most of the time with low $MSDV$, but as a result of her irregular driving, there are several situations that increase her average $MSDV$. Once again, the results fully agree with the feedback of the passengers, who rank drivers from D1 to D3, being D3 the most comfortable.

Finally, when vehicles are compared, the variable values determine that C2 is less comfortable in the longitudinal direction ($N_x$ and $MSDV_x$) but more comfortable if the lateral direction is considered ($N_y$ and $MSDV_y$). It can be argued that the fact that C2 is an old vehicle and very different to the ones the drivers are used to drive (especially with regards to the pedals and the gears), implies more longitudinal events. Moreover, they drove more carefully with C2, inducing less lateral events. Furthermore, kernel density estimation (Figure 9b) shows that driving with C2 is more steady, which is thought to be a consequence of the aforementioned conservative driving-style related to C2. It also depicts that the biggest differences between cars occur in sections S2 and S4, the ones which are the least comfortable ones.

As for the *z* axis, more vertical events are produced when passing through the potholes of the road when C1 is used, because although the suspensions of the C1 are newer than those of the C2, the speed during the journey is higher with C1. However, low-frequency power and, therefore, motion sickness are higher in that direction for C2. In any case, those effects do not seem to be meaningful according to the passenger feedback.

In [55], it was shown that the vehicle (in a simulator) presents a resonance frequency of 1.75 Hz which is a very similar conclusion to what it is stated in [56]. Furthermore, the authors of [57] performed a study about the effect of both the speed of the car and the road profile on the transmitability of the accelerations and proved that, in some cases, the biggest ones occur for 2 Hz. In the experimental campaign which has been carried out, this effect has been noted (Figure 10), but even though there are differences between car C1 and car C2, their power scale is quite low.

## 9. Conclusions and Future Work

The lack of comfort and motion sickness impact especially passengers; so, with a view to the autonomous car, a thorough analysis of the causes and effects influencing these sensations is essential. In this work, an experimental campaign has been designed as well as a method for the analysis of the data obtained which has proved to be useful to draw conclusions and, also, to analyze possible improvements for a more complex and therefore more decisive experimental campaign in order to understand how the future vehicle will be operating.

It has been concluded that the combination of the relative statistical analysis together with probability density statistical analysis and power spectral analysis allows us to obtain interesting conclusions about the effect that the road itself, the car and the driver have on the different variables that evaluate passenger comfort. Supporting this analysis with passenger feedback is essential to be able to compare quantitative results with the reality as perceived by passengers in order to evaluate the effectiveness of the method.

The results show that the principal factor in the lack of comfort on passengers is the type of the road itself. In fact, it has been observed that the spectrograms show certain patterns independently of the driver and the effect of the driver lies in the magnitude of the power at those moments, and therefore, this could be considered a second factor: the driving style. Besides, it has been demonstrated that the more comfortable is the road type, the more regular is the driving. So, in view of the autonomous car, the most decisive factor for minimizing the effect of motion sickness is a route planning system.

Moreover, a strong dependence of these patterns on the car type has been observed, which confirms the importance of vehicle dynamics in comfort (the main source of scientific articles in this field). However, results related with car type analysis are not decisive in this work, since it has been observed that the fact of not being familiar with the car has implied not driving so naturally with one of them.

In order to clarify the published information on which axes should be considered when evaluating motion sickness, the results show that contributions from longitudinal and lateral accelerations have a much greater effect than vertical ones (which are considered in the ISO standard).

Taking into account the obtained conclusions, a future experimental campaign is planned, in which the same journey will be considered but more drivers and cars will take part. Moreover, it is proposed that drivers familiarize themselves with the car before carrying out this campaign. On the other hand, and taking into account that the sensation of motion sickness is cumulative, the route will be divided in the same way, but, in addition, the experimental campaign must be carried out on each section separately. Likewise, a more complete questionnaire will be carried out and both the moments of discomfort (onboard) and the general feeling of comfort at the end of the trip will be recorded (offboard).

Finally, it has been proved that notwithstanding the variety of cars, drivers and sections, attending general discomfort, every situation of the experimental campaign is comfortable (small $A_v$). Therefore, this variable is useless to compare different situations. In order to also analyze changes in this variable, we propose a much more aggressive driving at certain times. However, taking into account that this could violate road safety regulations and, above all, could pose a danger to the passenger, we also propose to carry out an experimental campaign on a driving simulator such as the Virtual Development Center of the Automotive Intelligence Center [58].

**Author Contributions:** Conceptualization, project administration and funding acquisition, E.A. and I.d.C.; methodology, software, validation and formal analysis, O.M.-C., J.G.-Z. and J.A.R.; writing—original draft preparation and writing—review and editing, E.A., J.G.-Z. and J.A.R.; supervision, E.A., J.G.-Z., O.M.-C., I.d.C. and J.A.R. All authors have read and agreed to the published version of the manuscript.

**Funding:** This research was funded by Basque Government for partial support of this work through the project KK-2021/00123 *Autoeval* and the University of the Basque Country UPV/EHU under Grant GIU18/122.

**Institutional Review Board Statement:** Not applicable.

**Informed Consent Statement:** Not applicable.

**Data Availability Statement:** Not applicable.

**Conflicts of Interest:** The authors declare no conflict of interest.

## References

1. Bengler, K.; Dietmayer, K.; Farber, B.; Maurer, M.; Stiller, C.; Winner, H. Three Decades of Driver Assistance Systems: Review and Future Perspectives. *IEEE Intell. Transp. Syst. Mag.* **2014**, *6*, 6–22. [CrossRef]
2. Fafoutellis, P.; Mantouka, E.G.; Vlahogianni, E.I. Eco-Driving and Its Impacts on Fuel Efficiency: An Overview of Technologies and Data-Driven Methods. *Sustainability* **2021**, *13*, 226. [CrossRef]
3. Murray, L. Inside the future cars [Technology Driverless Cars]. *Eng. Technol.* **2017**, *12*, 60–62. [CrossRef]
4. Du, Y.; Liu, C.; Li, Y. Velocity Control Strategies to Improve Automated Vehicle Driving Comfort. *IEEE Intell. Transp. Syst.* **2018**, *10*, 8–18. [CrossRef]
5. Woolridge, E.; Chan-Pensley, J. Measuring User's Comfort in Autonomous Vehicles. Available online: https://humandrive.co.uk/wp-content/uploads/2020/07/HumanDrive-HF-Comfort-White-Paper-v1.1.pdf (accessed on 20 July 2020)
6. Forstberg, J. Ride Comfort and Motion Sickness in Tilting Trains. Ph.D. Thesis, Department of Vehicle Engineering, Royal Institute of Technology (KTH), Stockholm, Sweden, 2000.
7. Heibing, B.; Ersoy, M. *Chassis Handbook: Fundamentals, Driving Dynamics, Components, Mechatronics, Perspectives*; Viewg + Teubner Verlag: Wiesbaden, Germany, 2011.
8. Heinz. *Criteria for Ride Comfort in Buses and Coaches*; Transportation Research Board: Stockholm, Sweden, 1999.
9. Svensson, L.; Eriksson, J. Tuning for Ride Quality in Autonomous Vehicle: Application to Linear Quadratic Path Planning Algorithm. Ph.D. Thesis, Uppsala University, Uppsala, Sweden, 2015.
10. Javier, G.S. Generation of Ride Comfort Index. Ph.D. Thesis, Universidad Politécnica de Barcelona, Barcelona, Spain, 2014.
11. Griffin, M.J. Discomfort from feeling vehicle vibration, Vehicle System Dynamics. *Veh. Syst. Dyn.* **2007**, *45*, 679–698. [CrossRef]
12. Karjanto, J.; Wils, H.; Yusof, N.M.; Terken, J. Measuring the perception of comfort in acceleration variation using Eletro-Cardiogram and self-rating measurement for the passengers of the automated vehicle. *J. Eng. Sci. Technol.* **2022**, *17*, 18–196.
13. Daniela, P.; Parkhurst, G.; Shergold, I. Passenger comfort and trust on first-time use of a shared autonomous shuttle vehicle. *Transp. Res. Part C Emerg. Technol.* **2020**, *115*, 102604. [CrossRef]
14. Mühlbacher, D.; Tomzig, M.; Reinmüller, K.; Rittger, L. Methodological Considerations Concerning Motion Sickness Investigations during Automated Driving. *Information* **2020**, *11*, 265. [CrossRef]
15. Wang, C.; Zhao, X.; Fu, R.; Li, Z. Research on the Comfort of Vehicle Passengers Considering the Vehicle Motion State and Passenger Physiological Characteristics: Improving the Passenger Comfort of Autonomous Vehicles. *Int. J. Environ. Res. Public Health* **2020**, *17*, 6821. [CrossRef]
16. Nahvi, H.; Fouladi, M.H.; Jailani, M.; Nor, M. Evaluation of Whole-Body Vibration and Ride Comfort in a Passenger Car. *Int. J. Acoust. Vib.* **2009**, *14*, 143–149.
17. Marzbani, H.; Jazar, R.N. Smart flat ride tuning. In *Nonlinear Approaches in Engineering Applications 2*; Springer: Berlin/Heidelberg, Germany, 2014; pp. 3–36.
18. Lo, L.; Fard, M.; Subic, A.; Jazar, R. Structural dynamic characterization of a vehicle seat coupled with human occupant. *J. Sound Vib.* **2013**, *332*, 1141–1152. [CrossRef]
19. Múčka, P. Vibration dose value in passenger car and road roughness. *J. Transp. Eng. Part B Pavements* **2020**, *146*, 04020064. [CrossRef]
20. Martinez, C.M.; Heucke, M.; Wang, F.Y. Driving Style Recognition for Intelligent Vehicle Control and Advanced Driver Assitance: A survey. *IEEE Trans. Intell. Transp. Syst.* **2018**, *19*, 666–676. [CrossRef]
21. Naito, A.; Miyajima, C.; Nishino, T.; Kitaoka, N.; Takeda, K. Driver evaluation based on classification of rapid decelerating patterns. In Proceedings of the 2009 IEEE International Conference on Vehicular Electronics and Safety (ICVES), Pune, India, 1–12 November 2009; pp. 108–112.
22. Sathyanarayana, A.; Boyraz, P.; Hansen, J.H. Driver behavior analysis and route recognition by hidden Markov models. In Proceedings of the 2008 IEEE International Conference on Vehicular Electronics and Safety, Columbus, OH, USA, 22–24 September 2008; pp. 276–281.
23. Zylius, G. Investigation of Route-Independent Aggressive and Safe Driving Features Obtained from Accelerometer Signals. *IEEE Intell. Transp. Syst. Mag.* **2017**, *9*, 103–113. [CrossRef]
24. Elbanhawi, M.; Simic, M.; Jazar, R. In the Passenger Seat: Investigating Ride Comfort Measures in Autonomous Cars. *IEEE Intell. Transp. Syst. Mag.* **2015**, *7*, 4. [CrossRef]

25. Mata-Carballeira, Ó.; del Campo, I.; Asua, E. An eco-driving approach for ride comfort improvement. *IET Intell. Transp. Syst.* **2022**, *16*, 186–205. [CrossRef]
26. Hartwich, F.; Beggiato, M.; Krems, J. Driving Comfort, Enjoyment, and Acceptance of Automated Driving—Effects of Drivers' Age and Driving Style Familiarity. *Ergonomics* **2018**, *61*, 1017–1032. [CrossRef]
27. Bellem, H.; Schönenberg, T.; Krems, J.F.; Schrauf, M. Objective metrics of comfort: Developing a driving style for highly automated vehicles. *Transp. Res. Part F Traffic Psychol. Behav.* **2016**, *41*, 45–54. [CrossRef]
28. Qiao, X.; Ling, Z.; Li, Y.; Ren, Y.; Zhang, Z.; Ziwei, Z.; Qiu, L. Characterization of the Driving Style by State–Action Semantic Plane Based on the Bayesian Nonparametric Approach. *Appl. Sci.* **2021**, *11*, 7857. [CrossRef]
29. Smith, D.R.; Andrews, D.M.; Wawrow, P.T. Development and evaluation of the automotive seating discomfort questionnaire (ASDQ). *Int. J. Ind. Ergon.* **2006**, *36*, 141–149. [CrossRef]
30. Kolich, M. *Reliability and Validity of an Automobile Seat Comfort Survey*; Technical Report; SAE Technical Paper: Warrendale, PA, USA, 1999.
31. Lu, Z.; Li, S.; Felix, S.; Zhou, J.; Cheng, B. Driving comfort evaluation of passenger vehicles with natural language processing and improved AHP. *J. Tsinghua Univ. (Sci. Technol.)* **2016**, *56*, 137–143. [CrossRef]
32. Schoettle, B.; Sivak, M. *Public Opinion about Self-Driving Vehicles in China, India, Japan, the US, the UK and Australia*; UMTRI: Ann Arbor, MI, USA, 2014.
33. Kumara, V.; Saranb, V.; Guruguntla, V. Study of vibration dose value and discomfort due to whole body vibration exposure for a two wheeler drive. In Proceedings of the 1st International and 16th National Conference on Machines and Mechanisms (iNaCoMM2013), Roorkee, India, 18–20 December 2013; pp. 947–952.
34. Turner, M.; Griffin, M. Motion sickness in public road transport: The effect of driver, route and vehicle. *Ergonomics* **2000**, *42*, 1646–1664. [CrossRef] [PubMed]
35. Griffin, M.; Newman, M. An experimental study of low-frequency motion in cars. *Proc. Inst. Mech. Eng. Part D J. Automob. Eng.* **2004**, *218*, 1231–1238. [CrossRef]
36. Imaz, F.J.; Hurani, R.A.; Jaurena, J.F. *Estudio Del Índice de Confort del Servicio del Transporte Público de Pasajeros a Través de la Medición de Aceleraciones*; Repositorio de la Universidad Tecnologica Nacional: Santa Fe, Argentina, 2015.
37. Mechanical Vibration and Shock—Evaluation of Human Exposure to Whole-Body Vibration—Part 1: General Requirements. ISO 2631-1 International Organisation for Standarisation. Available online: https://www.iso.org/standard/7612.html (accessed on 20 July 2020).
38. Guide to Measurement and Evaluation Of human Exposure to Whole-Body Mechanical Vibration and Repeated Shock. BS 6841. Available online: https://standards.globalspec.com/std/978372/BS%206841 (accessed on 20 July 2020).
39. Griffin, M.J.; Newman, M.M. Visual field effects on motion sickness in cars. *Aviat. Space Environ. Med.* **2004**, *75*, 739–748.
40. Griffin, M.J.; Mills, K.L. Effect of frequency and direction of horizontal oscillation on motion sickness. *Aviat. Space Environ. Med.* **2002**, *73*, 537–543.
41. Donohew, B.E.; Griffin, M.J. Motion sickness: Effect of the frequency of lateral oscillation. *Aviat. Space Environ. Med.* **2004**, *75*, 649–656.
42. *Vibraciones y Choques Mecánicos. Evaluación de la Exposición Humana a Vibraciones de Cuerpo Entero. Parte 1: Requisitos Generales*; Standard; International Organization for Standardization: Geneva, Switzerland, 2008.
43. Castellanos, J.C.; Fruett, F. Embedded system to evaluate the passenger comfort in public transportation based on dynamical vehicle behavior with user's feedback. *Measurement* **2014**, *47*, 442–451. [CrossRef]
44. Kilinc, A.; Baybura, A. Determination of minimum horizontal curve radius usedin the design of transportation structures, depending on the limit value of comfortcriterion lateral jerk. In Proceedings of the Knowing to Manage the Territory, Protect Theenvironment, Evaluate the Cultural Heritage, Rome, Italy, 6–10 May 2012.
45. Cossalter, V.; Doria, A.; Garbin, S.; Lot, R. Frequency-domain method for evaluating the ride comfort of a motorcycle. *Veh. Syst. Dyn.* **2006**, *44*, 339–355. [CrossRef]
46. Sharma, R. Parametric analysis of rail vehicle parameters influencing ride behavior. *Int. J. Eng. Sci. Technol.* **2011**, *3*, 54–65. [CrossRef]
47. Yang, Y.; Ren, W.; Chen, L.; Jiang, M.; Yang, Y. Study on ride comfort of tractor with tandem suspension based on multi-body system dynamics. *Appl. Math. Model.* **2009**, *33*, 11–33. [CrossRef]
48. Stoica, P.; Moses, R.L. *Spectral Analysis of Signals*; Pearson Prentice Hall: Upper Saddle River, NJ, USA, 2005.
49. Solomon, O., Jr. PSD computations using Welch's method. *NASA STI/Recon. Tech. Rep. N* **1991**, *92*, 23584.
50. Cohen, L. *Time-Frequency Analysis*; Prentice Hall PTR: Englewood Cliffs, NJ, USA, 1995; Volume 778.
51. van Schagen, I.; Sagberg, F. The Potential Benefits of Naturalistic Driving for Road Safety Research: Theoretical and Empirical Considerations and Challenges for the Future. *Procedia-Soc. Behav. Sci.* **2012**, *48*, 692–701. [CrossRef]
52. Blatt, A.; Pierowicz, J.; Flanigan, M.; Lin, P.S.; Kourtellis, A.; Lee, C.; Jovanis, P.; Jenness, J.; Wilaby, M.; Campbell, J.; et al. *Naturalistic Driving Study: Field Data Collection*; Technical Report; Transportation Research Board: Washington, DC, USA, 2015.
53. Phyphox. Available online: https://phyphox.org/ (accessed on 20 July 2020).
54. Weglarczyk, S. Kernel density estimation and its application. *ITM Web Conf.* **2018**, *23*, 00037. [CrossRef]
55. Ibicek, T.; Thite, A. Quantification of Human Discomfort in a Vehicle Using a Four-Post Rig Excitation. *Low Freq. Noise Vib. Act. Control* **2012**, *31*, 29–42. [CrossRef]

56. Ibicek, T.; Thite, A. In Situ Measurement of Discomfort Curves for Seated Subjects in a Car on the Four-Post Rig. *Adv. Acoust. Vib.* **2014**, *2014*, 239178. [CrossRef]
57. Shakhlavi, S.J.; Marzbanrad, J.; Tavoosi, V. Various vehicle speeds and road profiles effects on transmitted accelerations analysis to human body segments using vehicle and biomechanical models. *Cogent Eng.* **2018**, *5*, 1461529. [CrossRef]
58. Automotive Intelligence Center Virtual Development Center. Available online: https://www.aicenter.eu/services/vdc (accessed on 14 October 2021).